\documentclass[sigconf]{acmart}

\usepackage[ruled]{algorithm2e}
\usepackage{subcaption}
\usepackage{multirow}
\newcommand{\norm}[1]{\left\lVert#1\right\rVert}

\usepackage{soul}

\def\BibTeX{{\rm B\kern-.05em{\sc i\kern-.025em b}\kern-.08emT\kern-.1667em\lower.7ex\hbox{E}\kern-.125emX}}
    
\begin{document}

%
\title{Aggregation Delayed Federated Learning}

%
\author{Ye Xue}
\affiliation{
  \institution{Northwestern University}
  \city{Evanston}
  \state{IL}
}
\email{ye.xue@u.northwestern.edu}

\author{Diego Klabjan}
\affiliation{
  \institution{Northwestern University}
  \city{Evanston}
  \state{IL}}
\email{d-klabjan@northwestern.edu}

\author{Yuan Luo}
\affiliation{
  \institution{Northwestern University}
  \city{Chicago}
  \state{IL}}
\email{yuan.luo@northwestern.edu}

%

%
\begin{abstract}
Federated learning is a distributed machine learning paradigm where multiple data owners (clients) collaboratively train one machine learning model while keeping data on their own devices. The heterogeneity of client datasets is one of the most important challenges of federated learning algorithms. Studies have found performance reduction with standard federated algorithms, such as FedAvg, on non-IID data. Many existing works on handling non-IID data adopt the same aggregation framework as FedAvg and focus on improving model updates either on the server side or on clients. In this work, we tackle this challenge in a different view by introducing redistribution rounds that delay the aggregation. We perform experiments on multiple tasks and show that the proposed framework significantly improves the performance on non-IID data.

\end{abstract}

%
%

%

%
\settopmatter{printfolios=true}
\maketitle

\section{Introduction}


As the amount of data generated by mobile devices increase explosively, followed by increasing privacy concerns of user data, researchers start seeking a solution to the dilemma of utilizing a large volume of user data while preserving the privacy of users. Federated learning is a machine learning paradigm that provides a solution to this dilemma. Under the coordination of a central server, a model is trained collaboratively by clients. To update the model, the server only collects a minimal amount of necessary information from clients but not their data \cite{mcmahan2017communication}. Federated learning has been drawing increasing interest in recent years and has been applied in many on-device prediction tasks \cite{hard2019federated, ramaswamy2019federated, bakopoulou2021fedpacket, chen2019federated}. The privacy promise of federated learning also makes it an appealing choice in healthcare applications \cite{silva2019federated, liu2019two, huang2019patient, huang2020loadaboost}. 

In federated learning, a global model is trained collaboratively on clients which are coordinated by a central server. Each round of training typically consists of four phases: an aggregation phase on the server, a local training phase on clients, and two communication (server-to-client and client-to-server) phases. The whole training process starts with a global model initialized on the server-side. In the server-to-client communication phase, a group of active clients is selected as training clients based on certain policy and the model is sent to them. Then, each client trains the model by calculating updates based on its own data stored on the local device. Stochastic gradient descent (SGD) is typically used to update local models during the local training phase. In the client-to-server communication phase, clients send their updated models back to the server, which then aggregates local models into a new global model in the aggregation phase. 

Different from traditional distributed learning, in federated learning, client's raw data are never collected by the central server. Instead, clients only send the updated model parameters to the server. In order to prevent potential information leakage from the local updated models, many privacy-preserving techniques, such as differential privacy, have been studied and applied to ensure privacy of model parameters \cite{geyer2018differentially, liu2019boosting, ghosh2019robust}. 

Since the server has no control of the client's data, this raises several challenges. First, data on a particular client are generated by a particular user, therefore, client data most likely are not distributed in a balanced and IID manner, which is usually an assumption in distributed learning. Second, the number of clients can be much larger than the number of samples on each client. This aggravates the issue of aggregation of non-IID clients. Third, clients are not always able to participate in training as user devices can be offline frequently or slow in communication. These challenges demand new methods different from existing algorithms designed for traditional distributed learning.

Our work focuses on mitigating the impact of non-IID client data distributions. Many existing works \cite{zhao2018federated,li2019gradient,xie2019asynchronous,li2018federated,sattler2019robust} adopt the FedAvg \cite{mcmahan2017communication} framework and applied various strategies to handle non-IID data. In this work, we propose a new framework of federated learning with delayed aggregations. We delay the aggregation of local models on the server by redistributing local models to clients multiple times. Compared with several state-of-the-art federated learning algorithms that handle non-IID data distributions, our framework demonstrates a good ability to mitigate the impact of the non-IID data distribution and yields the best performance on multiple datasets. We also propose an algorithm to select clients using importance sampling, incorporating which client further improves the performance of the algorithm. We implement our framework in Ray \cite{moritz2018ray} with code made public \footnote{Code is available at: \url{https://github.com/y-xue/RADFed}}. Our algorithm outperforms the best benchmark algorithm by $1.56\%$ on average across 9 non-IID datasets. On multiple datasets with a more challenging task, our algorithm demonstrates an improvement of roughly $3\%$ against the best comparison algorithm.

In order to better evaluate our algorithm's performance under non-IID settings, we propose a new method to generate non-IID data. Different from many existing sampling methods that focus only on sampling non-IID class distributions, the method can sample non-IID sizes of clients, class distributions and even feature distributions. Most importantly, it allows us to control the non-IID level for each of the three attributes separately. With this method, we can simulate different and specific non-IID settings. 

Along the way, we study the impact of localized and global data standardization. In global standardization, clients receive the mean and standard deviation of the global data distribution from the server and standardize local data using these statistics. Localized standardization is a procedure where each client standardizes the local data using its own statistics. Although global standardization is commonly used in federated learning studies, localized standardization is the only realistic choice when global statistics are not available. We observe a performance regression on federated algorithms when clients perform localized instead of global standardization, as expected. The proposed algorithms are robust to localized standardization scenarios, where we observe a larger improvement against comparison models than under global standardization.

In Section \ref{related}, we discuss related work. Proposed algorithms are described in Section \ref{meth}. The experimental setup, including dataset collection and generation, is described in Section \ref{exp}. Section \ref{res} discusses the computational results and the conclusions are drawn in Section \ref{conclusion}.

\section{Related Work} \label{related}




Many works have been done to tackle aforementioned challenges. Improving communication efficiency \cite{konevcny2016federated, mcmahan2017communication, luping2019cmfl} is one of the most important topics in federated learning as client devices are usually on slow and expensive connections. Performing sketched updates is a popular strategy. Kone{\v{c}}n{\`y} et al. \cite{konevcny2016federated} applied quantization and subsampling on the model update to compress it before sending it back to the server. Wang et al. \cite{luping2019cmfl} reduce communication by avoiding irrelevant updates from clients. Each client determines if its update is relevant enough by checking whether its local update aligns with the global tendency.

Despite the great success of FedAvg, researchers showed that the performance of FedAvg reduces significantly when local data are non-IID \cite{zhao2018federated}. Zhao et al. also proposed a strategy to mitigate non-IID data by sharing a subset of data between clients. The idea is to make the training data more IID through sharing. Many studies focuse on handling the non-IID issue in this direction \cite{jeong2018communication, yoshida2019hybrid, huang2020loadaboost}. Instead of sharing raw data, a generative adversarial network (GAN) was trained in \cite{jeong2018communication} to reproduce client data, which preserves privacy as no real data of clients is shared. 

Another category of studies improving federated learning on non-IID data adds constraints when updating the model. This can be done either on clients or on the server side. Sahu et al. proposed FedProx \cite{li2018federated}, which modified the loss function on the client side by adding a penalty to the weight difference between the local model and the global model. Xie et al. also added such penalty in their asynchronous federated learning algorithm \cite{xie2019asynchronous}. Sattler et al. proposed a communication-efficient federated learning framework to reduce communication costs by applying Top-k Sparsification \cite{sattler2019robust}. The sparsification restricted changes to only a small subset of the model's parameters and is shown to suffer the least from the non-IID data among existing model compression methods.

On the server side, Li et al. \cite{li2019gradient} applied momentum uniformly to the gradients of all clients to stabilize the training process under a non-IID scenario. However, collecting gradients from clients might require more frequent communications than collecting models from clients. Xie et al. \cite{xie2019asynchronous} proposed to update the global model by weighted averaging between new local updates and the old global model. Reddi et al. \cite{reddi2020adaptive} proposed adaptive federated learning algorithms, which treats the difference between the client's local update and the global model as pseudo-gradient and applied adaptive gradient descent algorithms to update the global model.

Our work focuses on handling non-IID data. Similar to \cite{li2018federated,reddi2020adaptive}, we modify the FedAvg algorithm to make it more robust on non-IID data. Different from existing works, we change the aggregation logic by introducing redistribution rounds which delay the aggregation. We also improve the client sampling process by incorporating the idea of importance sampling.

\section{Methodology} \label{meth}
One of the most common approaches to solve the optimization problem in federated learning is FedAvg \cite{mcmahan2017communication}. In each training round, the server sends the global model to a subset of randomly selected clients. The clients update their local model using SGD on their own data in parallel and send back the updated model to the server. The server then updates the global model by averaging local updates from clients. Consider a subset $\mathcal{K}$ of training clients, the aggregation at the $t$-th round is written as


\begin{displaymath}
    w_{t} \leftarrow \sum_{k \in \mathcal{K}} \frac{n_k}{n} w_{t}^{k} ,
\end{displaymath}

\noindent where $w^{k}_{t}$ is the updated model on client $k$, $n_k$ is the size of client $k$ and $n$ is the total size of clients. When data are identically distributed at clients, this aggregation works well since each local model is trained on a subset of data that is representative of the global distribution. It is identical to updating the global model in a centralized way. In non-IID cases, however, the client data can be highly skewed and it might not be a good idea to average the model weights trained on a highly skewed client with less skewed ones. The weighted averaging makes the aggregation even worse if a highly skewed client has a large number of samples, as the size of a client is also taken into account and a larger client has a bigger impact on aggregation. 

\subsection{Delayed Aggregation}

In order to make this aggregation work better in the non-IID setting, we have to answer the question: can each local model be trained on data that are representative of the global distribution at the time of aggregation? 

\begin{algorithm}
$K$ clients participate in training; $C$ is the fraction of clients participating in each training round; $T$ is the number of training iterations and $S$ is the number of redistributing iterations.  \\
\SetAlgoLined
\textbf{Server executes}:\\
initialize $w_{1}$\\
$m = max(C \cdot K, 1)$ \\
  \For{each round $t = 1,2,...,T$}{
    $w_{t}^{k} \leftarrow w_{t}$, for $k=1,2,...,m $\\
    $\bar{w}_t \leftarrow (w^{1}_{t},...,w^{m}_{t})$ \\
    \For{each redistributing iteration $s = 1,2,...,S$}{
      $U \leftarrow$ uniformly sample $m$ training clients \\
      \For{$i=1,2,...,m$}{
        $\bar{w}_{s+1}^{i} \leftarrow$ ClientUpdate($U_{i}, \bar{w}_{s}^{i}$)
      }
    }
    $w_{t+1} \leftarrow \frac{1}{m} \sum_{i=1}^m \bar{w}_{S+1}^{i}$ 
  }
\Return $w_{T+1}$ \\
\caption{RADFed}
\end{algorithm}

One of the core promises federated learning makes is that no client data is collected by the server, so we can not make data on each client be representative of the global distribution by rearranging client data. However, we can rearrange local models. If we train a model on all clients one by one, we end up with a model that is trained on all the data. This would be similar to standard epoch based training and thus very slow. Second, it would assume that each client is active when needed. Alternatively, we can select only a subset of clients to perform this strategy. Due to the fact that each client data can be skewed, the model may be trained on consecutive skewed mini-batches and thus might not be as good as the one trained in the centralized fashion, where data can be properly shuffled. Despite this, if each local model is trained on same data points at the time of aggregation regardless of the order of samples, we shall still expect a much more reliable aggregated model, compared with the case where each local model is only trained on data of a single client. 

Following this idea, we propose the Randomized Aggregation Delayed Federated learning algorithm (RADFed). We delay the aggregation by adding another training round to FedAvg. As shown in Algorithm 1, in the inner rounds, the server randomly sends local models back to clients again without performing aggregation. The server only aggregates local models at the end of the inner rounds. We call the inner training rounds the redistributing rounds. ClientUpdate($k$,$w$) trains the model of client $k$ with initial weights $w$.

With enough redistributions, all local models are expected to be trained on a similar number of samples. Therefore, we remove the sample size factor during aggregation and perform plain averaging over local models with equal weights. Because of this, the algorithm has another appealing property in terms of privacy-preserving in that the clients do not have to expose the size of their data. In many cases, the size of data can also be considered as sensitive information and exposing them may also cause privacy leakage. For example, it is more likely that a heavier user of a health-tracking app has a health problem.

In practice, it is possible that the number of active clients is different in each round. To apply our framework, a small subset of active clients can be selected to make sure the number of active clients is the same across redistributing rounds. In some extreme cases where too few clients are active during redistribution, there are multiple strategies to make the framework work, e.g., reducing the number of redistributing models accordingly, or counting the number of times a local model has been redistributed and scheduling the redistributing process to make sure local models are redistributed a similar number of times before aggregation.

\subsection{Importance Sampling}

Not all samples are equally important and so are clients, especially in federated learning where client data are usually non-IID. If data are not identically distributed on clients, why should we select training clients through a simple uniform random sampling? We hypothesize that focusing computation on good clients can help improve federated learning algorithms. Inspired by \cite{katharopoulos2018not}, we propose RADFed-IS that incorporates the idea of importance sampling into our aggregation delayed framework. In \cite{katharopoulos2018not} it is established that the optimal sampling probability is proportional to the square of the norm of the gradients.

The idea of importance sampling is to find a good mini-batch to train the model on in the next training step. A straightforward way of adopting this idea in our framework is to score the importance of all clients with respect to the current global model right after each aggregation and select the next set of clients to participate in training based on this score. However, collecting scores from all clients is usually not feasible in federated learning under the assumption that clients are not always active. Besides, it may increase the training time largely by adding an extra communication round to collect scores after each aggregation.

Instead, we score each client along with its local training. After local training, each client calculates the average square of the gradient norm of all mini-batches as its importance score and sends it back to the server along with the updated local model. The advantage of this strategy is that there is almost no extra burden added to the communication. Compared with the model itself, the size of an importance score can be neglected. However, the importance score calculated this way is no longer a good indicator of the importance of the client's data to the global model as each score is associated with a local model. In addition, a local model is not likely going to be trained on the same client in the next round because of the redistribution. Therefore, selecting clients based on this score might not be a good idea.

In order to solve this issue, we accumulate the importance scores for each client by averaging the scores calculated on all local models that have been trained on its local data. We expect that the accumulated score of a client becomes a good indicator of the importance of this client's data to all local models after accumulating over multiple rounds. 

The server accumulates importance score $p_{k}$ of client $k$ by taking a weighted average between the old score and the new one as 

\begin{displaymath}
    p_{k} \leftarrow (1-\alpha) p_{k} + \alpha p_{k}^{new}, 
\end{displaymath}

\noindent with a mixing hyper-parameter $\alpha \in (0,1)$. The detailed algorithm is shown in Algorithm 2.

\begin{algorithm}
\SetAlgoLined
\textbf{Server executes}:\\
initialize $w_{1}$\\
initialize $p_{k}$ for each training client $k$ \\
$m = max(C \cdot K, 1)$ \\
  \For{each round $t = 1,2,...,T$}{
    $w_{t}^{i} \leftarrow w_{t}$, for $i=1,2,...,m $\\
    $\bar{w}_t \leftarrow (w^{1}_{t},...,w^{m}_{t})$ \\
    \For{each redistributing iteration $s = 1,2,...,S$}{
      $U \leftarrow m$ clients sampled with probabilities $\propto p$ \\
      \For{$i=1,2,...,m$}{
        $\bar{w}_{s+1}^{i}, p_{U_{i}}^{new} \leftarrow$ ClientUpdate($U_{i}, \bar{w}_{s}^{i}$) \\
        $p_{U_{i}} \leftarrow$ $(1-\alpha) p_{U_{i}} + \alpha p_{U_{i}}^{new}$
      }
    }
    $w_{t+1} \leftarrow \frac{1}{m} \sum_{i=1}^m \bar{w}_{S+1}^{i}$ 
  }
\Return $w_{T+1}$ \\


\hrulefill \\
\textbf{ClientUpdate($k,w$)}:\\
  Client $k$ updates local model $w$ on local data $D$ \\
  $p = \frac{1}{|D|} \sum_{d \in D} || \nabla \ell_{d}(w)||^{2}_{2}$ \\
\Return $w$, $p$ to server
\caption{Importance Sampling in Federated Learning}
\end{algorithm}



\section{Experimental Setup} \label{exp}

In this work, we focus on evaluating the performance of federated learning algorithms in non-IID settings. Although a real-world non-IID dataset is ideal, datasets with an artificial partition are also very helpful in simulating different non-IID settings. Many studies create heterogeneous clients by manually sampling data on clients so that the class distribution is not identical across clients. In existing sampling methods, the sizes of clients are usually determined by class sampling. To the best of our knowledge, feature-imbalance has not been considered in prior works. 

In order to simulate non-IID settings with more control of the distribution of sizes, classes and features, we propose a sampling method where we can sample them independently with a different Dirichlet prior. It is not always the case that we can draw a desired number of samples to satisfy all these independently sampled distributions at the same time. Let us consider sampling non-IID sample sizes and classes as an example. A sampling solution for $T$ clients and $C$ classes is a $T \times C$ matrix where each entry denotes the number of samples of class $c$ on client $t$. By sampling sizes and classes separately, we specify each entry of the matrix, that is a total of $T \cdot C$ numbers. However, given the number of samples in each class in a dataset, we only need $T \cdot C - C$ entries to specify a solution. Therefore, we propose a Quadratic Programming (QP) method to find a random feasible sampling solution. 

\subsection{Partitioning of Heterogeneous Data}

\subsubsection{Non-IID over classes and sizes}
Let $C_{k}$ be the number of samples of class $k$ and $N$ be the total number of samples. Clearly, we have $N=\sum_{k} C_{k}$. Let $n \sim Dir(\mu)$ be the sizes of clients and $c_{t} \sim Dir(\lambda_{t})$ be the class distribution of client $t$. Let $\alpha_{tk}$ be the number of samples of client $t$ of class $k$. We want $\alpha_{tk} = c_{tk} n_{t} N$, given $n_{t}$ and $c_{tk}$. However, the dataset needs $\sum_{t} \alpha_{tk} = C_{k}$ and $\sum_{k} \alpha_{tk} = n_{t} N$, which they might not hold. Therefore, we find a feasible solution for $\alpha$ by solving 
\begin{equation}
    \min_{\alpha \ge 0} \sum_{t,k} (\alpha_{tk} - c_{tk} n_{t} N)^2
\end{equation}

\noindent subject to: 

\begin{displaymath}
\begin{split}
    \sum_{k} \alpha_{tk} &= n_{t} N, \forall \text{ client } t, \\
    \sum_{t} \alpha_{tk} &= C_{k}, \forall \text{ class } k , \\
\end{split}
\end{displaymath}
\noindent which is a convex QP.

\subsubsection{Non-IID over features, classes and sizes}
Using the similar idea of sampling classes and sizes, we also sample categorical features in a non-IID manner. We sample category distribution $f^{j}_{t} \sim$ Dir($\theta^j_t$) of feature $j$ with $d_{j}$ categories on client $t$. We consider classes as a feature that are sampled separately. Let $U$ be a set of all possible combinations of categories in the dataset and $B_{u}$ be the number of samples that fall into configuration $u \in U$. The first element of $u$ corresponds to classes. Now let $\alpha_{tu}$ be the number of samples on client $t$ with configuration $u$. Then we find a feasible solution through 

\begin{equation}
\begin{split}
    \min_{\alpha \ge 0} \sum_{t}  [&\sum_{k=1}^{K} (\sum_{u \in U, u_1 = k} \alpha_{tu} - c_{tk} n_t N)^2 \\ &+  \sum_{j=1}^{M} \sum_{i=1}^{d_{j}} (\sum_{u \in U, u_{j+1}=i} \alpha_{tu} - f^{j}_{ti} n_t N)^2]
\end{split}
\end{equation}

\noindent subject to: 

\begin{displaymath}
\begin{split}
\sum_{u \in U} \alpha_{tu} &= n_t N, \forall \text{ client } t, \\
\sum_{t} \alpha_{tu} &= B_{u}, \forall \text{ configuration } u \in U . \\
\end{split}
\end{displaymath}
\noindent Here $M$ is the number of categorical features. If a feature is non-categorical by nature, we can create buckets that correspond to categories.

                    
\subsubsection{A random solution}
The above QPs may have many optimal solutions but we want a random one. We generate a random solution by modifying values at the ``4 vertices of a random rectangle,'' in a way that the modified values still satisfy our constraints in (1) or (2), see details in Algorithm 3. A step size $\xi$ is used to control the modification. The algorithm has two phases. In the first phase, we find a suboptimal solution by randomly modifying values. Then, in the second phase, starting from the suboptimal solution, we continue modifying and record the best random solution we find. 

\begin{algorithm}

Input: a feasible solution $A=\{\alpha_{tk}\}_{tk}$ from QP \\
\For{$p = 1,2,...,P$}{
    $A \leftarrow$ RandomizeSolution($A$) // a burn-in period \\
}
$h \leftarrow \infty$ \\
\For{$q = 1,2,...,Q$}{
    $A \leftarrow$ RandomizeSolution($A$) \\
    $L(A) \leftarrow$ calculate loss from (1) or (2) \\
    \If{$L(A) < h$}{
        $\bar{A} \leftarrow A$ \\
        $h \leftarrow L(A)$
    }
}
\Return $\bar{A}$ \\
\hrulefill \\
\textbf{RandomizeSolution($A$)}:\\
    $(i,j), (\bar{i},\bar{j}) \leftarrow$ indices of two randomly selected entries of $A$ \\
    $\varepsilon \leftarrow \text{uniform}(0, \min\{A_{i,j}, A_{\bar{i},\bar{j}}, \xi\})$ \\
    \For{\text{Each position} $(m,n)$}{
    $A^{new}_{m,n} \leftarrow \left\{
                            \begin{array}{ll}
                            A_{m,n} - \varepsilon, m=i, n=j \\
                            A_{m,n} - \varepsilon, m=\bar{i}, n=\bar{j}\\
                            A_{m,n} + \varepsilon, m=i, n=\bar{j}\\
                            A_{m,n} + \varepsilon, m=\bar{i}, n=j\\
                            A_{i,j}, \text{otherwise}\\
                            \end{array}
                        \right. $ \\
    }
    \Return $A^{new}$ \\
\caption{Random QP solution}
\end{algorithm}

\begin{table}
  \centering
  \caption{Statistics of datasets (number of samples of clients)}
  \begin{tabular}{lrrrrrr}
    \toprule
    Dataset&Min&Max&Mean&Stdev&C-score\\
    \midrule
    Cifar10    & 2 & 2,850 & 600 & 605.85 & 1.286 \\ 
    Shakespeare    & 3 & 41,305 & 3,616 & 6,808.44 & 0.266 \\ 
    COVCLS    & 110 & 33,300 & 4,920 & 5,110.28 & 0.794 \\ 
    COVFEAT    & 372 & 17,328 & 4,920 & 3,237.06 & 0.682 \\  
    MNIST $\lambda=1$    & 3 & 3,365 & 700 & 667.12 & 0.696 \\ 
    MNIST $\lambda=0.1$    & 11 & 3,327 & 700 & 658.89 & 1.293 \\ 
    eICU    & 108 & 5,683 & 901 & 925.50 & 0.060  \\
    
  \bottomrule
\end{tabular}
\label{tab:data}
\end{table}

\begin{figure*}[htp]

\centering
\includegraphics[width=.33\textwidth]{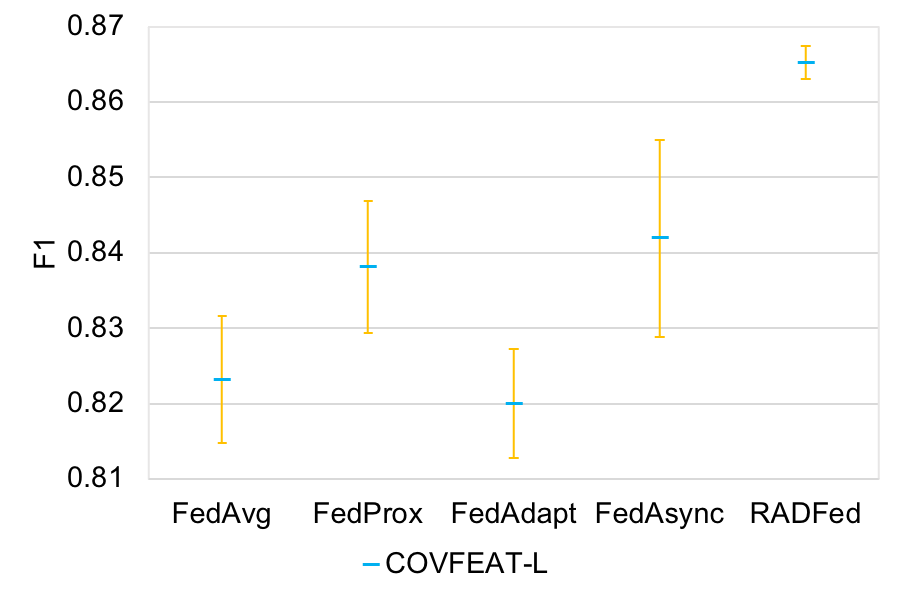}\hfill
\includegraphics[width=.33\textwidth]{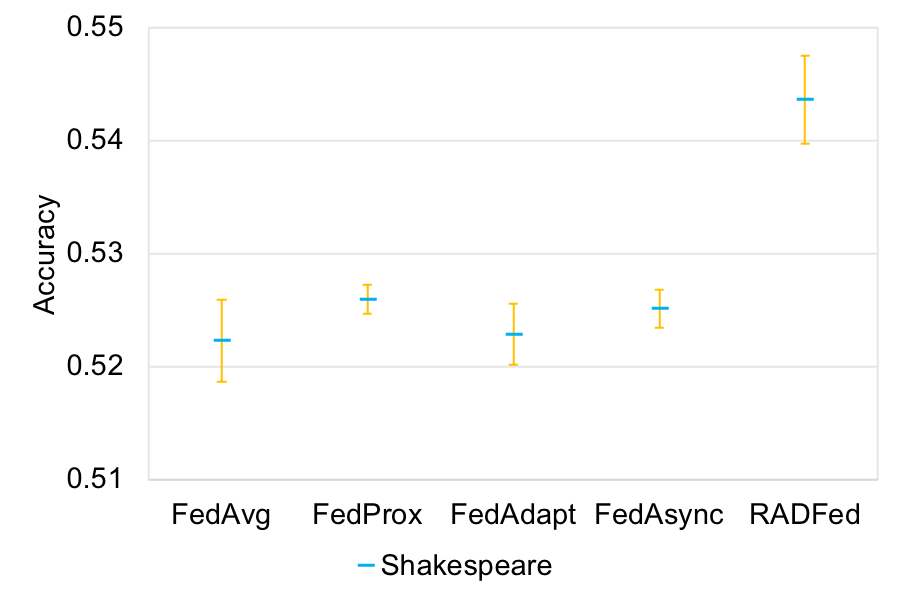}\hfill
\includegraphics[width=.33\textwidth]{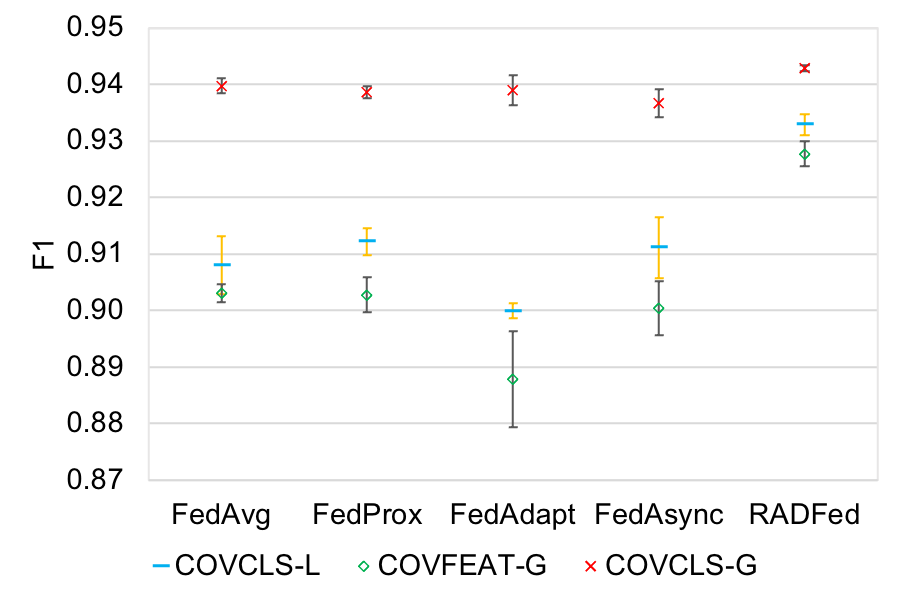}
\caption{Test performance comparison on the Covertype and Shakespeare datasets. Multiple runs are performed with different seeds on the most representative fold, defined as the one with the closest performance gap to the average of all folds. The performance gap is the difference in the test performance between RADFed and FedAvg.}
\label{fig:error_bar}

\end{figure*}

\begin{table*}
  \centering
  \caption{Average test performance of 5-fold cross-validation: \% accuracy for the MNIST, Cifar10 and Shakespeare dataset; F1 score ($\times$100) for all Covertype datasets; and Area Under the Receiver Operating Characteristic Curve (AUC) ($\times$100) for the eICU dataset. The best values are shown in bold. The absolute scores are reported for FedAsync and the \% relative performance difference against FedAsync is shown for other algorithms.}
  \begin{tabular}{l*7r}
    \toprule 
  
    Dataset&FedAvg&FedProx&FedAsync&FedAdapt&RADFed&RADFed-IS\\
    \midrule 
    Cifar10  & $-$2.07& $-$1.24& (84.63) & $-$1.69& $+$0.61& \bfseries $+$0.90  \\
    Shakespeare  & $-$0.75& $-$0.42& (52.10)& $-$0.21& $+$3.26& \bfseries $+$3.72  \\ 
    COVFEAT$-$G   & $-$0.34& $+$0.02& (87.91)& $-$1.79& \bfseries $+$2.88& $+$2.24  \\
    COVFEAT$-$L   & $-$0.92& $+$0.36& (79.52)& $-$2.40& \bfseries $+$4.05& $+$3.32  \\
    COVCLS$-$G   & $+$0.20& $+$0.23& (93.68)& $-$0.05& $+$0.36& \bfseries $+$0.44  \\
    COVCLS$-$L   & $-$0.06& $+$0.32& (90.67)& $-$0.29& \bfseries $+$2.22& $+$0.03  \\
    MNIST $\lambda=1$  & $+$0.05& $+$0.19& (97.24)& $+$0.15& $+$0.22& \bfseries $+$0.28  \\ 
    MNIST $\lambda=0.1$  & $+$0.17& $+$0.32& (96.79)& $+$0.27& $+$0.40& \bfseries $+$0.45  \\ 
    eICU   & $-$0.11& $-$0.11& \bfseries (92.31)& $-$0.05& $\mathbf{+0.00}$& $-$0.03  \\
    \midrule 
    AVG & $-0.42$ & $-0.04$ & - & $-0.67$ & $+1.56$ & $+1.26$ \\
   \bottomrule
\end{tabular}
\label{tab:test_scores}
\end{table*}

\subsection{Datasets and Models}

For all datasets that are partitioned by (1) or (2), we set $\mu=1$. These datasets have reasonably large variations in client sizes, see Table \ref{tab:data}. We use the same $\lambda=0.1$ for all clients, on all datasets but MNIST, where we experiment on different values of $\lambda$. For feature sampling, we set $\theta = 0.1$ for all clients and features. In QP, we use $P = 10^5$, $Q = 5 \cdot 10^5$ and $\xi=0.002$. The impact of $C$, $B$ (the mini-batch size) and $E$ (the number of local training epochs) is well studied and thus we do not focus on experimenting on various settings of these variables. We set $C=0.1$, which is shown to be a generally good setting that balances the performance and the convergence speed \cite{mcmahan2017communication}. The mini-batch size $B$ is set to 10 and 16 for MNIST and Cifar10, respectively, considering that clients on these datasets do not have many samples. On other datasets, $B$ is set to 256. We set $E=10$ for MNIST to make the task more challenging and set $E=1$ for the other datasets. Besides these general federated learning hyper-parameters as mentioned above, each particular algorithm has its own hyper-parameters. RADFed has one more hyper-parameter, the number of redistribution rounds $S$, than FedAvg. RADFed-IS adds another hyper-parameter, the mixing weight $\alpha$. We tune hyper-parameters specified by each federated learning algorithm using grid search on validation clients. Table \ref{tab:params} lists the hyper-parameter values of proposed algorithms used in our experiments.

\begin{table*}
  \centering
  \caption{Hyper-parameters in proposed algorithms}
  \begin{tabular}{l*7r}
    \toprule 
  
    param&COVCLS (-L \& -G) & COVFEAT (-L \& -G) & MNIST $\lambda=1$&MNIST $\lambda=0.1$&Cifar10&Shakespeare&eICU\\
    \midrule 
    $S$ (RADFed)            & 22  & 20  & 22  & 15  & 15  & 15  & 80 \\
    $S$ (RADFed-IS)         & 22  & 20  & 22  & 15  & 100 & 100 & 80 \\
    $\alpha$ (RADFed-IS)    & 0.9 & 0.9 & 0.9 & 0.9 & 0.9 & 0.8 & 0.9 \\
   \bottomrule
\end{tabular}
\label{tab:params}
\end{table*}

\subsubsection{Covertype} Covertype is a large structured dataset for forest cover type prediction from the UCI KDD archive \cite{ucikdd}. It consists of 10 numerical features and 2 categorical features with 7 imbalanced classes. Since our goal is to evaluate our method's performance on non-IID data, we do not want to consider other data quality problems such as high class imbalance at the same time. Therefore, in our experiments, we only focus on predicting the two largest classes. We partition the data into 100 clients. The number of training clients ($K$) is $60$. The number of validation and test clients are 20 each. The splitting of clients is discussed in Section \ref{eval}. Same sizes are used for the MNIST and Cifar10 datasets. We train a fully connected neural network with 2 hidden layers with 64 neurons each.

\begin{figure*}[htp]

\centering
\includegraphics[width=.3\textwidth]{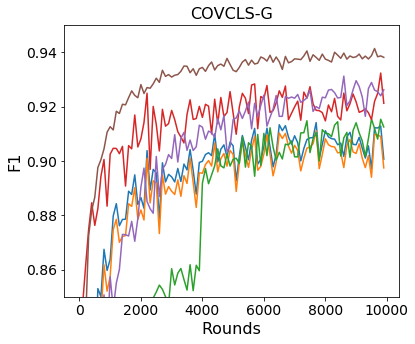}\hfill
\includegraphics[width=.3\textwidth]{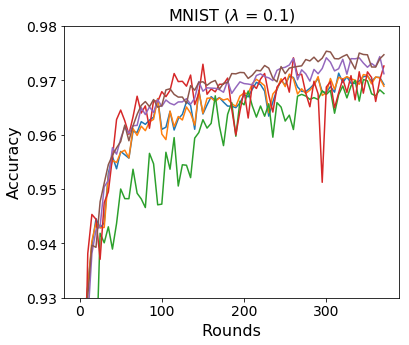}\hfill
\includegraphics[width=.3\textwidth]{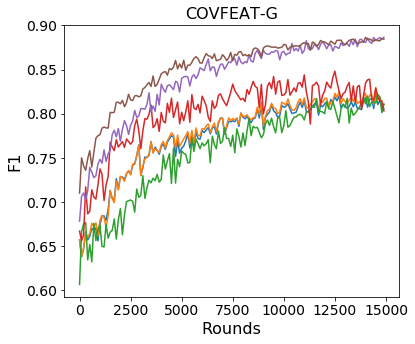}\hfill
\includegraphics[width=.3\textwidth]{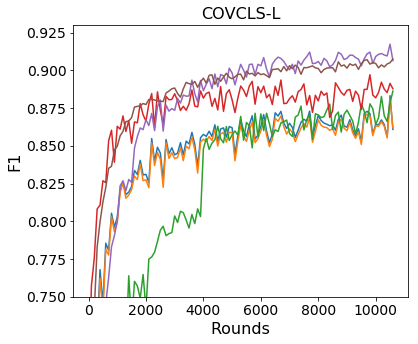}\hfill
\includegraphics[width=.3\textwidth]{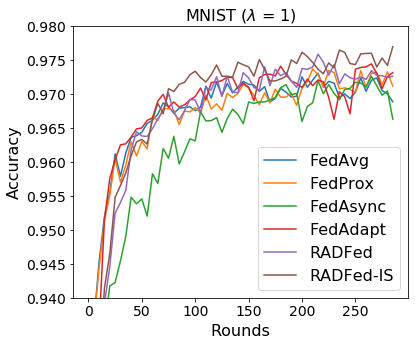}\hfill
\includegraphics[width=.3\textwidth]{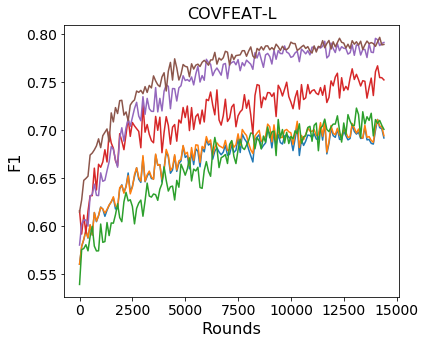}

\caption{Validation performance comparison on the Covertype and MNIST datasets. The F1 score is used on all Covertype datasets and accuracy is reported on the MNIST datasets. Curves are smoothed by taking the average over evenly spaced intervals for better visualization. The intervals are chosen differently considering that validation frequencies are different. The intervals are set to 100 for the Covertype datasets and 5 for the MNIST datasets.}
\label{fig:val_curves_1}

\end{figure*}

\begin{figure*}[htp]

\centering
\includegraphics[width=.3\textwidth]{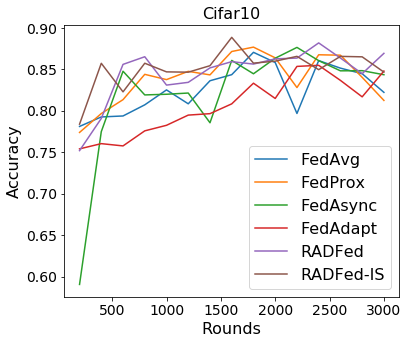}\hfill
\includegraphics[width=.31\textwidth]{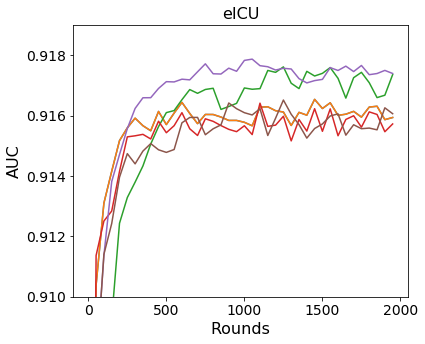}\hfill
\includegraphics[width=.3\textwidth]{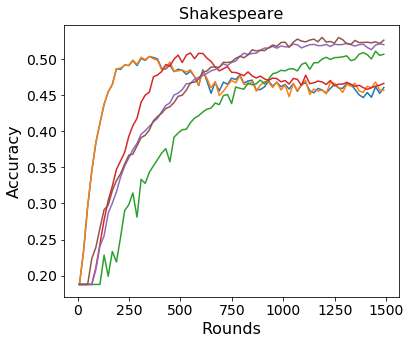}

\caption{Validation performance comparison. Accuracy is reported on the Cifar10 and Shakespeare datasets. AUC is used on the eICU dataset. Similar to Figure \ref{fig:val_curves_1}, curves are smoothed with intervals 1, 2 and 50 for the Cifar10, Shakespeare and eICU datasets, respectively.}
\label{fig:val_curves_2}

\end{figure*}

We create two types of datasets, one (COVCLS) with classes and client sizes sampled non-identically based on (1) and the other (COVFEAT) with also features sampled non-identically thus using (2). For both datasets, we set $\lambda=0.1$ for all clients, and set $\theta=0.1$ for the COVFEAT dataset. All the datasets that follow are created based on (1). 

On the Covertype datasets, we also study the impact of localized and global data standardization. The difference is whether to use global statistics of all clients' data to standardize client local data or to let each client perform standardization with its own statistics. On COVFEAT-G and COVCLS-G, we perform global standardization, while on COVFEAT-L and COVCLS-L, localized standardization is used. When comparing our algorithm with benchmarks on other datasets, we use global standardization to be consistent with the original papers.

\subsubsection{MNIST} MNIST \cite{lecun2010mnist} consists of images of digits with 10 classes. We sample 100 clients with classes and sizes non-identically distributed. We study how data heterogeneity impacts the performance of federated learning algorithms by creating two datasets with $\lambda=1$ and $\lambda=0.1$, respectively. A dataset generated with the larger $\lambda$ has a lower heterogeneity in class distributions. We build a fully connected neural network same as \cite{mcmahan2017communication}. 

\subsubsection{Cifar10} Cifar10 \cite{krizhevsky2009learning} images are partitioned into 100 clients with classes ($\lambda=0.1$) and sizes ($\mu=1$) non identically distributed. We use pre-trained MobileNetV2 \cite{sandler2018mobilenetv2} as the model and train a subset of layers from the last bottleneck convolution layer to the classification layer. 

\subsubsection{Shakespeare} This dataset is a language modeling dataset built from \textit{The Complete Works of William Shakespeare} \cite{mcmahan2017communication}. We use the same data as \cite{li2018federated} but partition samples by speaking roles. Each speaking role corresponds to one client. In total, the dataset consists of 143 clients. The number of training, validation and test clients are $85,29$ and $29$, respectively. The task is to predict the next character given a sequence of 80 characters. We train a 2 layer long short-term memory (LSTM) classifier with an 8-dimensional embedding layer. 

\subsubsection{eICU} eICU is a large multi-center critical care database made available by Philips Healthcare \cite{pollard2018eicu}. We predict the in-hospital mortality using variables underlying the Acute Physiology Age Chronic Health Evaluation (APACHE) predictions \footnote{The full variable list and descriptions are available at \url{https://eicu-crd.mit.edu/eicutables/apachepredvar} and \url{https://eicu-crd.mit.edu/eicutables/apacheapsvar/}.}. To avoid a potential sampling bias, we focus on mid to large hospitals with more than 100 admissions and exclude those associated with a high mortality rate (greater than 20\%). Each hospital corresponds to a client. The dataset contains 164 clients. The number of training, validation and test clients are $98,33$ and $33$, respectively. We train a logistic regression model with L2 regularization.


\begin{figure*}[htp]

\centering
\includegraphics[width=.3\textwidth]{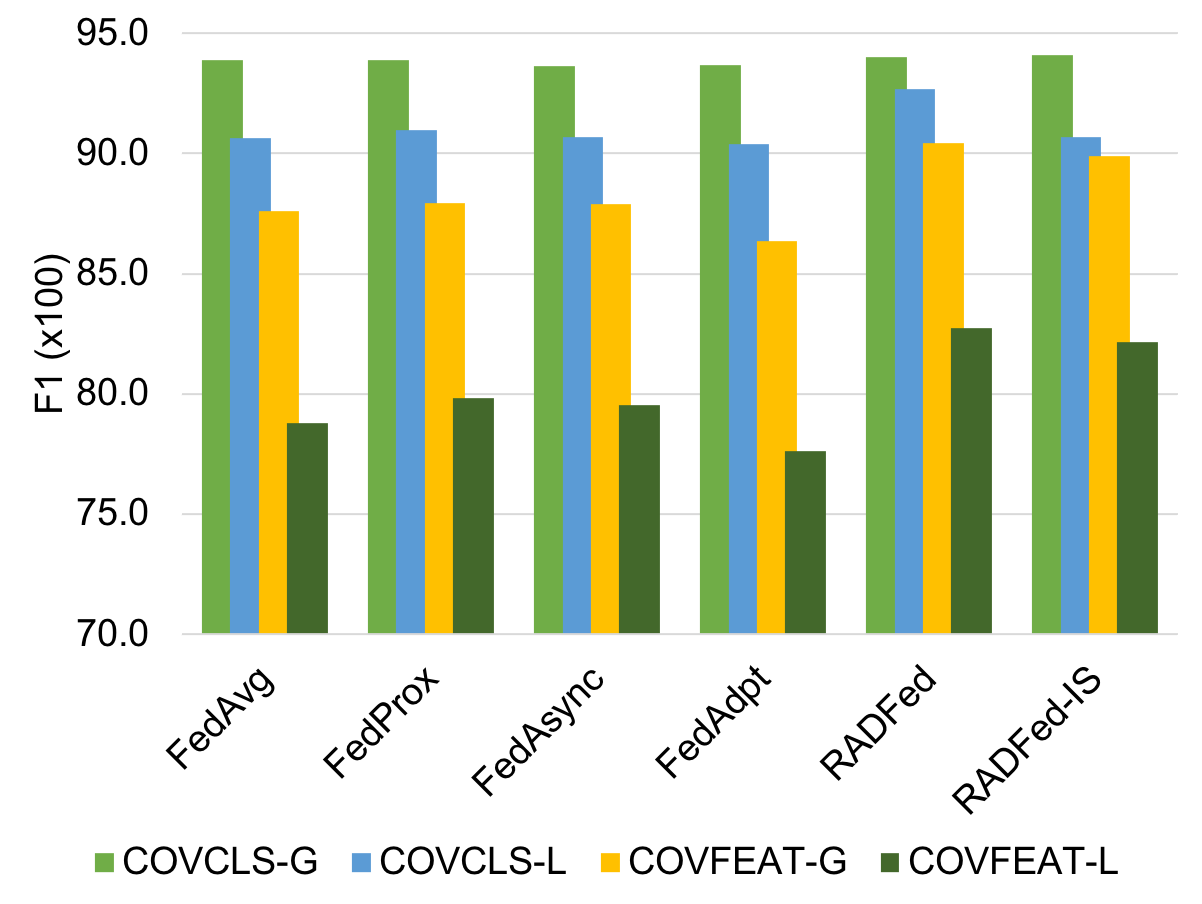}
\includegraphics[width=.3\textwidth]{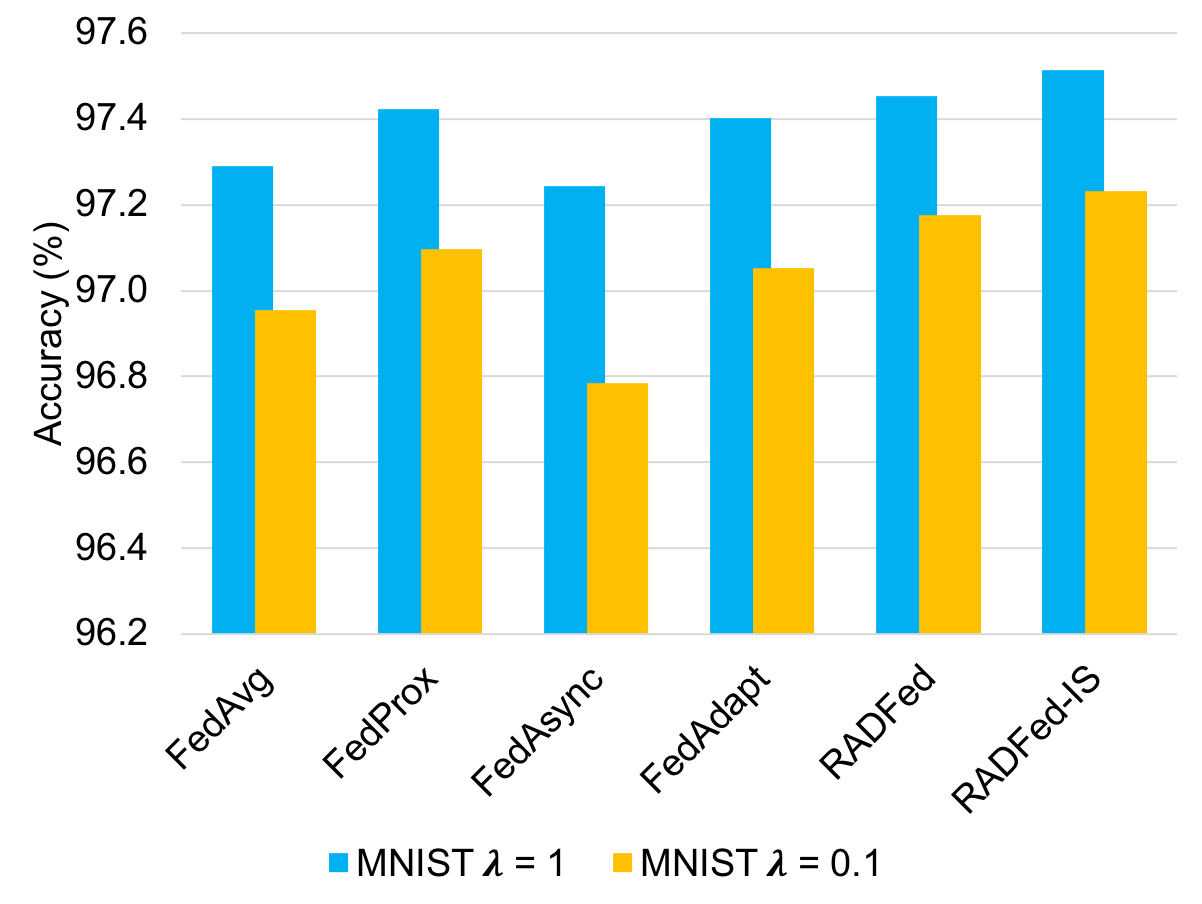}

\caption{Test performance comparison on datasets with different levels of heterogeneity}
\label{fig:hetero}

\end{figure*}

\subsection{Evaluation Setup} \label{eval}

We compare the performance of our methods (RADFed and RADFed-IS) with FedAvg \cite{mcmahan2017communication}, FedProx \cite{li2018federated}, the adaptive federated operation method (FedAdapt) \cite{reddi2020adaptive} and the asynchronous federated optimization method (FedAsync) \cite{xie2019asynchronous} on multiple tasks. FedAvg is probably the most popular and commonly used federated algorithm and the others are the state-of-the-art federated learning algorithms that handle non-IID data distributions. 

Different from synchronous methods, FedAsync has to deal with the staleness of updates from clients. The staleness of a client's update is defined as the timestamp difference between a client's update and the server's model. The performance of FedAsync suffers from large staleness. In order to mitigate the impact of staleness on training, the new global model is updated as a weighted average between the old global model and the client's local update. In addition, the authors show that decaying the mixing weights as a function of staleness helps to fight against large staleness. Despite these efforts, the impact of staleness on FedAsync's performance is not completely eliminated. 

In order to make a fairer comparison between asynchronous and synchronous methods, we have to choose a reasonable value for staleness. We simulate the FedAsync's training procedure and find maximum staleness where the average number of clients running in parallel per round is the same as in the synchronous methods. In other words, we compare the performance of FedAsync and synchronous methods under the same level of parallelism on average. 

To our best knowledge, there is no gold standard for evaluating federated algorithms. Generally, there are 3 ways to split the data into training and test sets: splitting all data globally \cite{sattler2019robust,li2019gradient,huang2019patient}, splitting each client's local data \cite{ghosh2019robust,reddi2020adaptive,bakopoulou2021fedpacket} and splitting clients into training/test groups \cite{huang2020loadaboost,hard2019federated,ramaswamy2019federated}. In this work, we adopt the last strategy by assuming no local data can be collected by the server and the server can not manipulate the client's local data. Additionally, we perform 5-fold cross-validation with the by-client splits in order to reduce the selection bias, which might be aggravated by the non-IID client distributions. We split all clients into 5 sets. One by one, a set is selected as the test set. For the remaining sets, one by one, a set is selected as the validation set and the others are used as the training set.


\section{Results} \label{res}

We run each algorithm 3 times with different seeds on each of the 5 folds and report the average performance over the 15 runs in Table \ref{tab:test_scores}. On average, RADFed and RADFed-IS offer an improvement over the best benchmark, FedAsync, by 1.56\% and 1.26\%, respectively. On the MNIST datasets and the eICU dataset, all algorithms achieve a close performance. On other datasets, the best of RADFed and RADFed-IS is significantly better than FedAsync (p < 0.05 under the Wilcoxon signed-rank test \cite{wilcoxon1992individual}). Under some difficult settings, which we discuss later, our framework offers a substantial improvement over all comparison algorithms on multiple datasets. The best of RADFed and RADFed-IS outperforms the best comparison algorithm by $3.72\%$ on Shakespeare, $2.88\%$ on COVFEAT-G, $4.05\%$ on COVFEAT-L and $2.22\%$ on COVCLS-L. Figure \ref{fig:error_bar} shows that RADFed is quite stable across different seeds and confirms its significant improvement on these datasets.

In Figures \ref{fig:val_curves_1} and \ref{fig:val_curves_2}, we compare the validation curves. With delayed aggregations, RADFed and RADFed-IS stabilize the training by demonstrating a smaller variation in validation scores than the algorithms that adapt the FedAvg framework. In general, our algorithms achieve the maximum validation score at a similar number of training rounds as other algorithms. On Shakespeare, our algorithms peak much later than FedAvg. This is due to a large learning rate used in FedAvg, where a relatively larger learning rate yields a better result, although the model gets overfitted quicker than using a lower learning rate. It does not imply that delaying aggregations also delays convergence. In fact, on Shakespeare, aggregations in RADFed are delayed with 15 redistribution rounds and the number in RADFed-IS is 100. We observe a similar convergence behavior. 

\subsection{Heterogeneity}

We create datasets with various levels of heterogeneity to evaluate whether our model is effective and robust under different heterogeneous settings. In order to compare between manually partitioned datasets and naturally partitioned ones, we introduce the class non-IID score (C-score), which is defined as $\frac{1}{K} \sum_{k=1}^{K} \sum_{c=1}^{C} |r^{k}_{c} - R_{c}|$, where $r^{i}_{c}$ is the ratio of class $c$ on client $k$ and $R_{c}$ is the ratio of class $c$ in all data. This score measures the difference between client's class ratios and the global class ratios. The C-score of each dataset is shown in Table \ref{tab:data}. 

The MNIST dataset is partitioned with $\lambda=1$ and $\lambda=0.1$. The MNIST $\lambda=0.1$ dataset is expected to have a higher heterogeneity of class distributions than the other due to a smaller value of $\lambda$. Its C-score is also higher than in MNIST $\lambda=1$. All algorithms perform worse on MNIST $\lambda=0.1$ than MNIST $\lambda=1$, as shown in Figure \ref{fig:hetero}. The RADFed-IS algorithm performs the best on both datasets and yields the smallest performance regression when class heterogeneity increases. 

The COVCLS and COVFEAT datasets are partitioned with the same value of $\mu$ and $\lambda$, so they have a similar level of heterogeneity with respect to client sizes and classes. Their C-scores are also similar. However, since we also introduce heterogeneity on feature distributions in the COVFEAT datasets, they should have a severer issue on non-IID data distribution than COVCLS datasets. Therefore, as shown in Figure \ref{fig:hetero}, all algorithms show a lower performance on COVFEAT datasets, no matter which standardization method is used. Similar to the results on the MNIST datasets, our algorithms suffer the least when data heterogeneity increases.

\subsection{Standardization}

With global standardization, RADFed-IS achieves a close performance as RADFed on both COVFEAT-G and COVCLS-G datasets. RADFed outperforms FedAvg by 0.17\% and 3.24\% on COVCLS-G and COVFEAT-G, respectively. 

With localized standardization, we observe a performance regression on all federated learning algorithms, as shown in Figure \ref{fig:hetero}. However, RADFed demonstrates a good ability in handling localized standardization by offering a larger performance improvement over FedAvg on both COVCLS-L (2.3\%) and COVFEAT-L (5.0\%). 

Interestingly, RADFed outperforms RADFed-IS on both COVCLS-L and COVFEAT-L, which implies that it is more challenging for RADFed-IS to determine which clients are better under localized standardization. Therefore, we recommend RADFed-IS when it is possible to perform global standardization and RADFed under localized standardization.




\subsection{Model Complexity}

On all Covertype and MNIST datasets, we train fully connected neural networks with hidden layers. While on Cifar10 and Shakespeare datasets, we test our algorithms with deeper networks (i.e., Convolutional Neural Networks and Recurrent Neural Networks) on unstructured tasks. RADFed-IS performs the best on both datasets. On Cifar10, it improves RADFed by 0.28\% and outperforms the best comparison algorithm and FedAvg by 0.90\% and 3.03\%, respectively. Different from Cifar10, Covertype and MNIST datasets, the Shakespeare dataset is not manually partitioned through our sampling algorithm. Instead, the data are partitioned naturally by speaking roles. As a result, the Shakespeare dataset has a lower C-score than these three datasets. Our RADFed-IS algorithm improves RADFed by 0.45\% and surpasses all other comparison methods by at least 3.26\%.

Besides deep models, we train a logistic regression model on the eICU dataset. As shown in Table \ref{tab:test_scores}, all federated algorithms achieve similar AUC scores. The scores are all close to the centralized model's AUC of 0.924. The eICU dataset is also not partitioned through the sampling algorithm and its C-score is much smaller than the scores of all other datasets. Despite small heterogeneity of the dataset, RADFed still offers an improvement over FedAvg, modest but significant ($p=0.001$ under the Wilcoxon signed-rank test \cite{wilcoxon1992individual} with 15 runs).

\subsection{Divergence on Delayed Updates}

\subsubsection{Divergence from the centralized model}

In studying the non-IID challenge in federated learning, the weight divergence has been used to explain the performance reduction, which as shown in \cite{zhao2018federated} can be attributed to the divergence. The weight \textbf{D}ivergence between the \textbf{C}entralized and federated models ($DC$) measures the difference of the global weights of federated training relative to those of centralized training. It is defined as

\begin{equation}
    DC(t) = \frac{||\mathbf{w}^t_{FL}||-||\mathbf{w}^t_{C}||}{||\mathbf{w}^t_{C}||} ,
\end{equation}

\noindent where $\mathbf{w}^t_{FL}$ are the weights of the global model in federated training at the $t$-th round and $\mathbf{w}^t_{C}$ are the centralized weights.

\begin{figure}[htbp]
    \centering
    \captionsetup{justification=centering}
    
    \begin{subfigure}{0.5\columnwidth}
        \includegraphics[width=\textwidth]{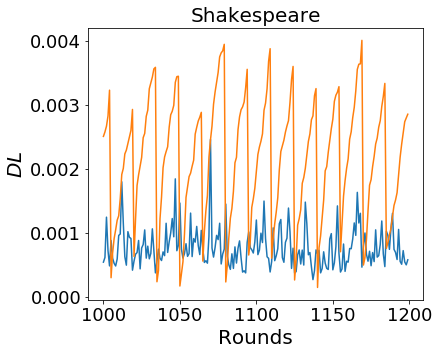}
        \caption{}
        \label{fig:dl_sp}
    \end{subfigure}\hfill
    \begin{subfigure}{0.47\columnwidth}
        \includegraphics[width=\textwidth]{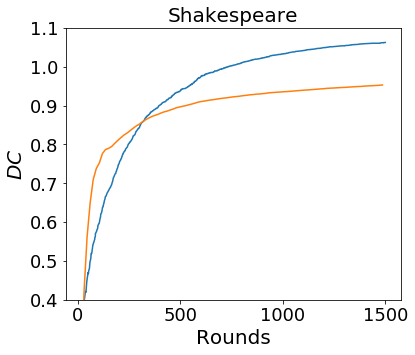}
        \caption{}
        \label{fig:dc_sp}
    \end{subfigure}
    
    \begin{subfigure}{0.47\columnwidth}
        \includegraphics[width=\textwidth]{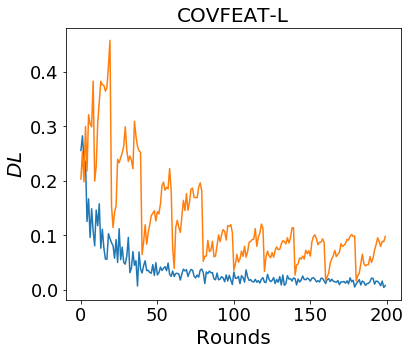}
        \caption{}
        \label{fig:dl_cov}
    \end{subfigure}\hfill
    \begin{subfigure}{0.5\columnwidth}
        \includegraphics[width=\textwidth]{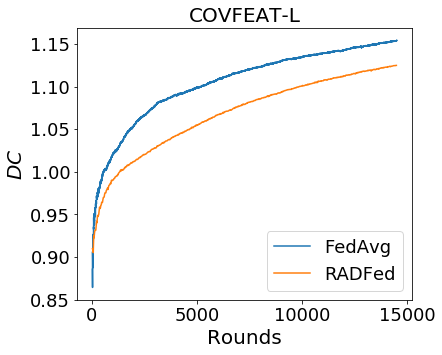}
        \caption{}
        \label{fig:dc_cov}
    \end{subfigure}

    \caption{Weight divergence}\label{fig:divergence}
\end{figure}

To visualize the weight divergence $DC$, we train a centralized model and a federated model side by side. Both models start with the same weight initialization. In each round, the same data are used in training. The difference is that in centralized training we collect data from clients and update the model using combined data. The divergence from the centralized model is expected due to the distance between the client data distribution and the population distribution. As shown in Figures \ref{fig:dc_sp} and \ref{fig:dc_cov}, RADFed algorithm demonstrates a smaller weight divergence than FedAvg. It indicates that the aggregated weights of our algorithm are less impacted by the skewness of the data and are closer to the weights trained on data under the population distribution.

\subsubsection{Divergence between local models}

Another type of weight divergence is the \textbf{D}ivergence between clients' \textbf{L}ocal updates ($DL$) before each aggregation. For a set of clients $\mathcal{K}=\{1,2,...,K\}$, the divergence is defined as

\begin{equation}
    DL(\mathbf{w}^t_1, \mathbf{w}^t_2, ... \mathbf{w}^t_K) = \binom{K}{2}^{-1} \sum_{i,j \in \mathcal{K}; i<j} (1-\frac{\mathbf{w}^t_i \cdot \mathbf{w}^t_j}{\norm{\mathbf{w}^t_i} \norm{\mathbf{w}^t_j}}) ,
\end{equation}

\noindent where $\mathbf{w}^t_i$ is the local update from client $i$ in the $t$-th round. A positive correlation between $DL$ and federated learning performance is observed in \cite{kim2019federated}. The study is based on the FedAvg framework that is different from ours. Although the same correlation might not hold when comparing different frameworks, it helps visualize how our algorithm behaves.

During training of our algorithm, we observe a periodical trajectory of $DL$, Figure \ref{fig:divergence}. In the first round after each aggregation, the divergence is the smallest. As the aggregation being delayed for more rounds, the divergence keeps increasing until the next aggregation. The divergence in FedAvg vibrates around the lowest values of our algorithm. Figures \ref{fig:dl_sp} and \ref{fig:dl_cov} show the weight divergence of local updates on Shakespeare and COVFEAT-L datasets. 

The increasing $DL$ does not indicate any deficiency of our framework. It might be due to the nature of the redistribution of local models. For example, in a non-IID setting where each client has one class of data, training may start with clients of different classes and yield large divergence between local models. In the next redistribution round, the divergent local models are trained again on client data of different classes. The divergence accumulates as the redistribution continues. FedAvg, however, results in a smaller $DL$ because it performs aggregation after each local training and divergence is not accumulated.





\section{Conclusion} \label{conclusion}

In this work, we propose a new training framework with delayed aggregation to handle the well-known non-IID issue in federated learning. We demonstrate that our framework offers a substantial improvement over the FedAvg framework and outperforms several state-of-the-art federated learning algorithms. Moreover, we incorporate importance sampling in our framework and further improve the framework on multiple datasets. 

Along the way, we also discuss the following topics in federated learning: the splitting of training and test sets, localized and global standardization, and weight divergence on different frameworks. Experiments show that federated learning algorithms suffer from localized standardization. The proposed framework demonstrates a good ability in handling localized standardization. However, the importance sampling version does not offer further improvement under localized standardization.

In addition, we propose a sampling algorithm to generate non-IID datasets. It offers the choice for a desired non-IID level on client sizes, classes and features separately, thus providing researchers with more flexibility and control about simulating different non-IID settings. We also introduce the C-score to quantify the level of heterogeneity of non-IID datasets and demonstrate the robustness of proposed algorithms on datasets with various C-scores.

%

%
\bibliographystyle{ACM-Reference-Format}
\bibliography{sample-base}

%


\end{document}